\documentclass[runningheads]{llncs}
\usepackage{graphicx}
\usepackage{tikz}
\usepackage{comment}
\usepackage{amsmath,amssymb} % define this before the line numbering.
\usepackage{color}
\usepackage{commath}
\DeclareMathOperator*{\argmax}{arg\,max}
\DeclareMathOperator*{\argmin}{arg\,min}
\begin{document}
% \linenumbers
\pagestyle{headings}
\mainmatter
\def\ECCVSubNumber{100}  % Insert your submission number here

\title{Blind Image Deblurring: a Review} % Replace with your title

\titlerunning{Blind Image Deblurring: a Review}
%\author{First Author\inst{1}\orcidID{0000-1111-2222-3333} \and
%Second Author\inst{2,3}\orcidID{1111-2222-3333-4444} \and
%Third Author\inst{3}\orcidID{2222--3333-4444-5555}}
%
\author{Zhengrong Xue}
\authorrunning{Zhengrong Xue}

\institute{Shanghai Jiao Tong University}
%******************
\maketitle

\begin{abstract}
In this paper, we first formulate the blind image deblurring problem and explain why it is challenging. Next, we bring some psychological and cognitive studies on the way our human vision system deblurs. Then, relying on several previous reviews, we discuss the topic of metrics and datasets, which is non-trivial to blind deblurring, and introduce some typical optimization-based methods and learning-based methods. Finally, we review the highlights of 
%two classical methods (Wiener~\cite{gonzalez2008digital} and Richardson-Lucy~\cite{richardson1972bayesian,lucy1974iterative}), 
the theoretical analysis part of Levin \textit{et al.}~\cite{levin2009understanding}, which could answer some very fundamental questions about blind deblurring.
\end{abstract}

\section{Problem Formulation}
\label{section_problem}
Image blur can be caused by a variety of reasons including camera shake~\cite{fergus2006removing}, object motion~\cite{jia2007single} or defocus~\cite{zhou2011coded}. Camera shake blur comes from the camera vibration when the shutter is pressed, which affects all the pixels in an image. Object motion blur is detectable when we attempt to capture some moving object at high speeds. While the foreground object is blurry, the background is usually sharp. Defocus blur refers to blurry scenes outside the depth of field. Depth of field (DOF) is the range covered by all objects in a scene that appear acceptably sharp in an image.

Mathematically, the blurring degradation process can be formalized as
\begin{align}
    y=k\otimes x+n,
    \label{formulation}
\end{align}
which says the observed blurry image $y$ is a convolution of an unknown sharp image $x$ and unknown kernel $k$ plus some noise $n$. The noise $n$ is usually assumed to be Gaussian noise~\cite{levin2009understanding}, Poisson noise~\cite{ma2013dictionary} or impulse noise~\cite{cai2010fast}. In signal processing, the convolution kernel is also known as point spread function (PSF). For simplicity, most studies introduced in this paper presume $k$ is spatially invariant so that the blind image deblurring problem is reduced to a standard blind deconvolution problem. But in most real-world cases, $k$ is unfortunately to be spatially variant~\cite{levin2009understanding}, which makes the problem much trickier. What's worse, as is stated in \cite{lai2016comparative}, the camera motion has 6 degrees of freedom (3 translations and 3 rotations) while the convolution model only considers 2D translations parallel to the image plane.

But now let's just consider the simplified blind deconvolution problem. At first sight, one might think it quite trivial once he/she has learned some knowledge about Fourier transform. However, blind deconvolution is one of the most ill-posed problems in signal processing. In fact, even non-blind deconvolution is not trivial at all.

\begin{figure}[t]
\centering
\includegraphics[height=6.5cm]{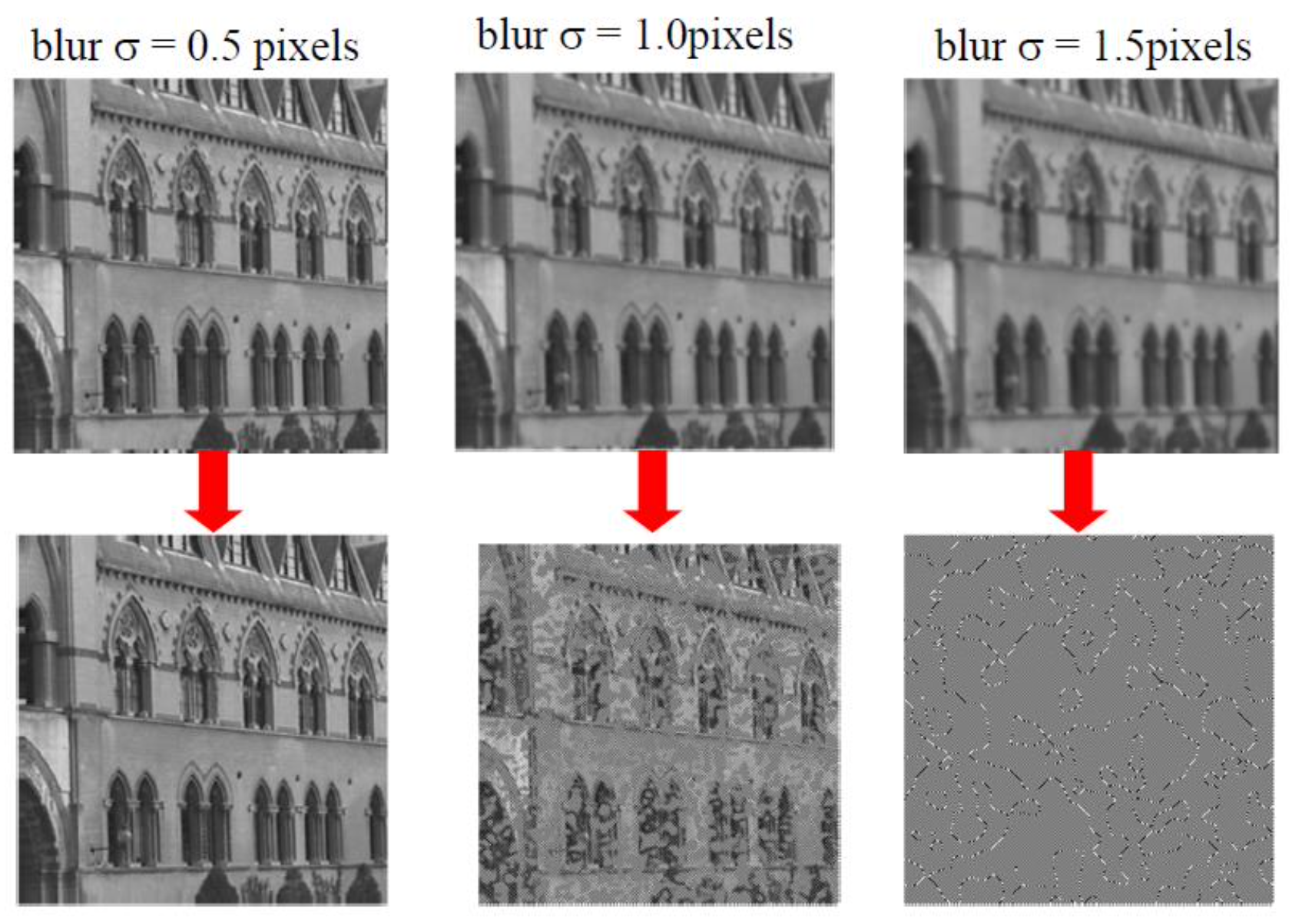}
\caption{The results of deblurring directly with an inverse filter. It can be observed that larger kernel leads to worse results. This is because larger kernel in spacial domain is smaller in frequency domain, which intensifies the noise amplification effect. Source: Lecture notes of Digital Signal and Image Processing instructed by Yuye Lin.}
\label{naive_inverse}
\end{figure}

Suppose now we perfectly know what $k$ is like and we would like to adopt a naive inverse filtering method to recover the blurry image. We have:
\begin{align}
    \tilde{x}=\mathcal{F}^{-1}\left \{\frac{1}{\mathcal{F}\left \{k\right \}}\mathcal{F}\left \{y\right \}\right \}=\mathcal{F}^{-1}\left \{\mathcal{F}\left \{x\right \}+\frac{1}{\mathcal{F}\left \{k\right \}}\mathcal{F}\left \{n\right \}\right \},
\end{align}
where $\mathcal{F}\left \{\cdot\right \}$ and $\mathcal{F}^{-1}\left \{\cdot\right \}$ denote Fourier transform and inverse Fourier transform respectively. Since $k$ is a Gaussian signal, its Fourier transform $\mathcal{F}\left \{k\right \}$ should still follows a Gaussian distribution, which means most of its energy concentrates in low frequencies. If $n$ is assumed to be white noise, then it has constant intensity across the spectrum. Consequently, the noise is dramatically amplified at high frequencies, and the latent sharp image is covered up with noise. Fig.~\ref{naive_inverse} shows some examples of unsatisfactory results.

With no knowledge of the kernel at all, blind deconvolution is apparently more difficult to tackle. We say it is a severely ill-posed problem, mainly because there are infinite sets of $(x,k)$ satisfying Eq.~\ref{formulation}. For example, it could be a no-blur explanation, \textit{i.e.}, $k$ is the delta (identity) kernel and $x=y$.

\section{Psychological and Cognitive Studies} 
Before introducing artificial intelligence, why not first examine the intelligence of human vision?

All grand events like the opening ceremony of the Olympics are accompanied with dazzling fireworks, and fireworks in the dark seem to leave behind a long tail to human vision. This effect which cartoonists or digital displays take advantage of to make still pictures look like moving is known as visible persistence\cite{coltheart1980iconic}. Since photographs taken of moving objects would be blurry at slower shutter speeds, the human vision system whose `shutter speed' is around 1/8 second\cite{burr1981temporal} should probably suffer the same problem. Nevertheless, real-life experience tells us most moving objects look sharp for human beings. This indicates our vision system may be equipped with some kind of motion-deblurring mechanism. 

\begin{figure}[t]
\centering
\includegraphics[height=6cm]{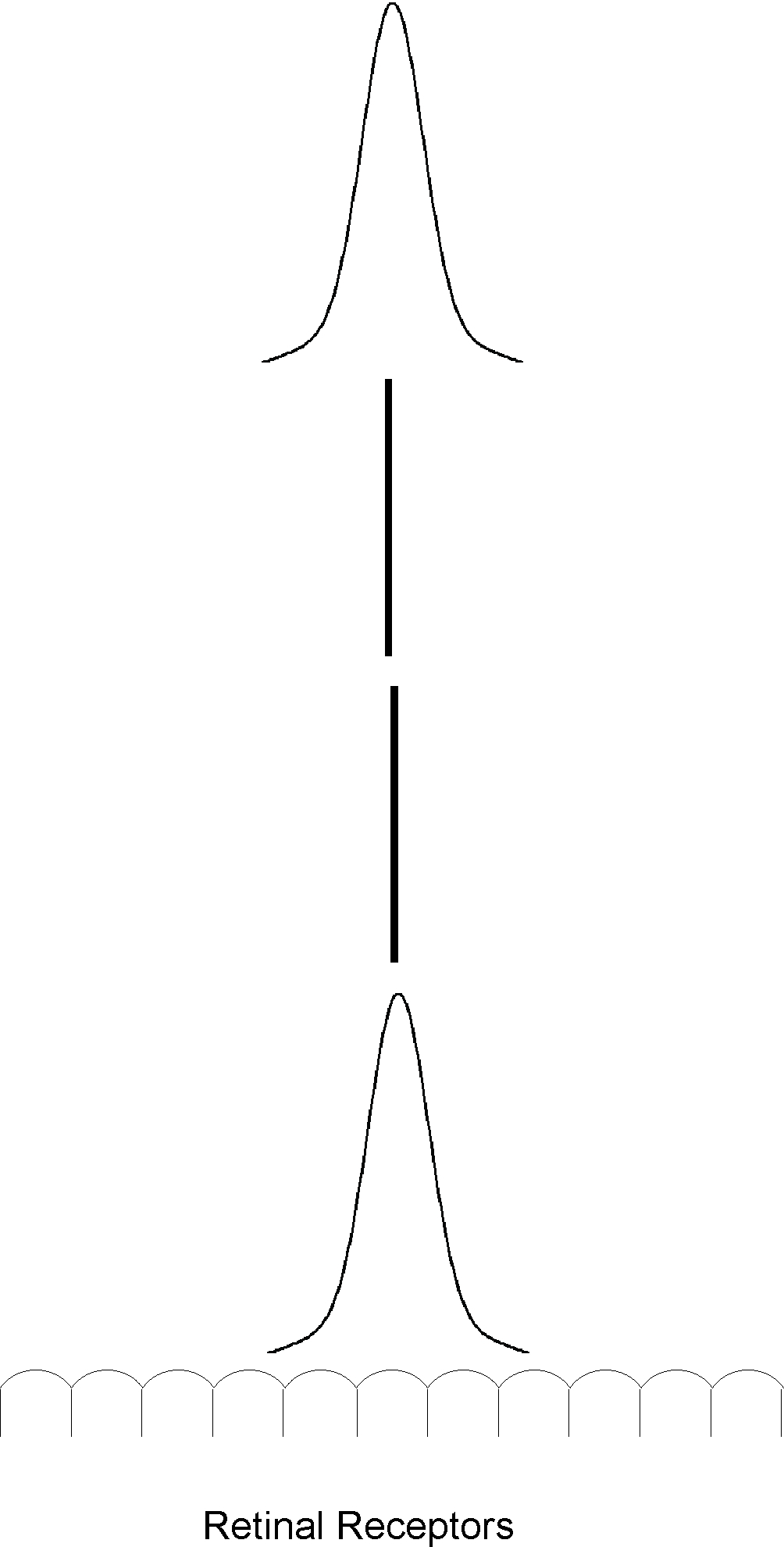}
\includegraphics[height=5.5cm]{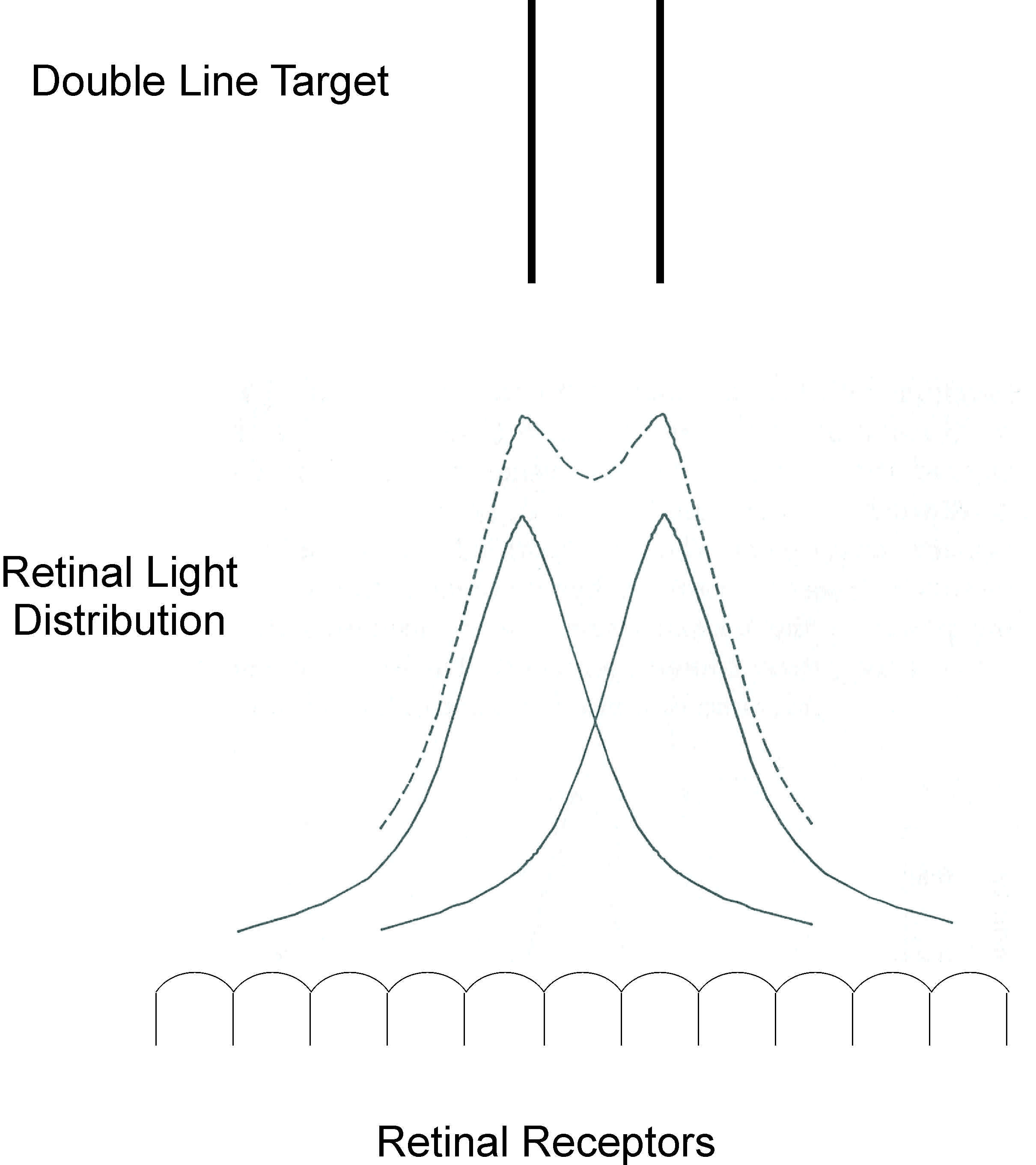}
\caption{(Left) In hyperacuity, the criterion is not one or two but precision of target locations relative to one another, shown here an offset pair of vernier lines. (Right) A standard visual acuity paradigm, where two targets have to be far enough apart so that their combined light energy pattern (dotted lines) produces stimulation in members of the receptor array that allows the differentiation between one and two. Source: Scholarpedia.}
\label{hyperacuity}
\end{figure}

Psychologists and cognitive scientists have raised different assumptions\cite{anderson1987shifter,burr1986seeing} on motion-deblurring mechanism in human vision. Evidence that may demonstrate the existence of the mechanism is one vernier hyperacuity experiment\cite{westheimer1975visual} (See Fig.~\ref{hyperacuity} for some intuition), in which the observer must detect the direction of offset between two lines with bordering ends. Experimental results show that human's capability of discriminating this very precise spatial pattern is unaffected by certain types of motion. However, another kind of hyperacuity conducted by a different study\cite{morgan1989motion} shows that discrimination of the distance between two parallel lines is interfered with motion. This argues against a general deblurring mechanism, such as a neural network \textit{shifter circuit} proposed by Anderson \textit{et al.}\cite{anderson1987shifter}.

To the best of our knowledge, there is no generally accepted biological or psychological theory yet to illustrate the mechanism behind motion-deblurring ability of our vision system.

\section{Existing Approaches and Related Work}
\subsection{Previous Reviews}
 Despite the challenges mentioned in the first two sections, numerous approaches have been proposed by the computer vision and signal processing community to tackle the challenging blind deblurring problem. The most efficient way to have a rough grasp of these approaches is to rely on previous reviews.

 An informative review~\cite{kundur1996blind} published in 1996 introduces a great number of classical blind image deconvolution methods. In this review, various methods are categorized into zero sheet separation methods~\cite{lane1987automatic}, a priori blur identification methods~\cite{cannon1976blind} and ARMA parameter estimation methods~\cite{lagendijk1990identification}. We will not directly cover those methods in this paper, but the techniques that classical methods adopted, like maximum a posteriori probability (MAP) estimate or expectation–maximization (EM) algorithms, are frequently leveraged in contemporary methods.
 
Levin \textit{et al.} from the famous MIT CSAIL gave a thoughtful technical report~\cite{levin2011understanding} in 2011. The report theoretically explains why naive MAP approach on both the latent sharp image and the kernel ($\mathrm{MAP}_{x,k}$) often fails, and why MAP on the kernel only ($\mathrm{MAP}_{k}$) and then non-blind deconvolution is a more reasonable choice. While most of the arguments in this paper are quite inspiring and convincing, some assumptions seem to be so imprecise that may damage the correctness of the theory. What's more, the theory is not general enough to explain all the methods leveraging MAP estimate, including some most successful ones~\cite{cho2009fast,xu2010two}. Nevertheless, mathematical analysis explaining why some algorithms work and why others do not is always most valuable in computer science. Thus, we will introduce the highlights of~\cite{levin2011understanding} in Section~\ref{section_theory_bayesian}.
 
 Another review~\cite{wang2014recent} in 2014 groups then-existing methods into five categories: Bayesian inference framework, variational methods, sparse representation-based methods, homography-based modeling, and region-based methods. While the criteria that the paper adopts to classify various methods is sometimes a little bit confusing, this paper does cover a very wide range of approaches.

\begin{figure}[b]
\centering
\includegraphics[height=2.6cm]{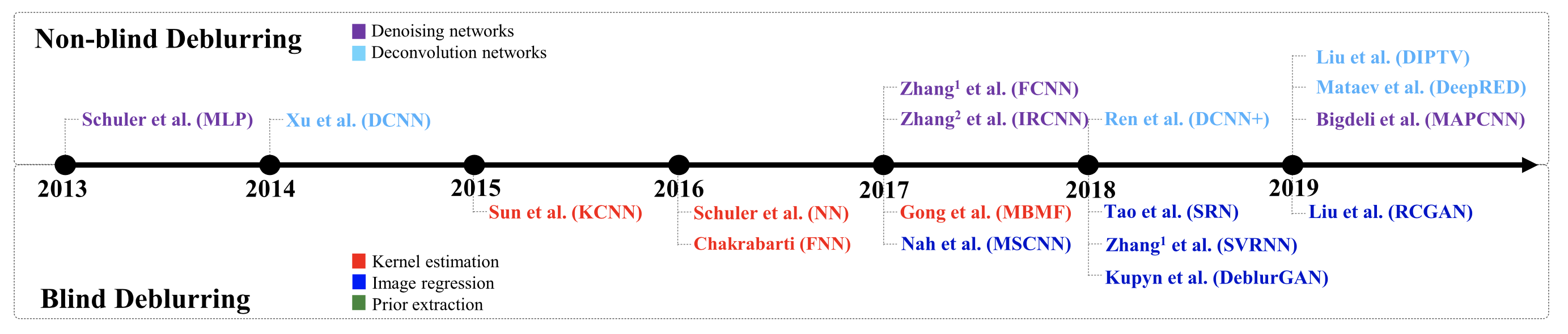}
\caption{Milestones in non-blind and blind deblurring based on deep learning. In blind deblurring, early methods (marked in red) take advantage of neural networks to estimate the kernel. After Nah \textit{et al.}~\cite{nah2017deep} is proposed, direct image-to-image regression methods (marked in blue) are developed. Source: Koh \textit{et al.}~\cite{koh2021single}}
\label{milestone}
\end{figure}

In the era of neural network, deep learning has revolutionized almost every topic in computer vision. Deblurring is a classical low-level vision problem. Unsurprisingly, many learning-based deblurring methods have been raised in recent years. A very recent review~\cite{koh2021single} illustrates how learning-based deblurring methods are developed (See Fig.~\ref{milestone}), and evaluates the performance of different approaches. This paper also presents insightful opinions on the main challenges in deblurring with neural networks. Firstly, the dataset that is essential for all learning-based methods is not easy to prepare. Synthesizing technique~\cite{levin2009understanding} cannot simulate non-uniform blur, realistic blurry images~\cite{kohler2012recording,lai2016comparative} are often too expensive to acquire, while averaging methods~\cite{hyun2015generalized,nah2017deep} are available only for dynamic scenes, thus lacking diversity. For more information, see Section~\ref{metrics_dataset}. Secondly, an extensive blur has a long-range spatial dependency. The receptive field of CNNs are not big enough, while RNNs or LSTM are often unstable and require too many computational resources. Thirdly, if the training set contains both small and large blurs, optimal parameters for small and large blurs might be different. Since the parameter space is continuous, the parameters might take intermediate values during the learning process~\cite{lehtinen2018noise2noise}, which is optimal for the whole set, but very bad for some individual images.

\subsection{Metrics and Datasets}
\label{metrics_dataset}
Unlike many other computer vision tasks such as classification or segmentation whose effects can be easily quantified and compared, it is rather difficult to judge how well an algorithm deblur an image. Of course, there are metrics like peak signal-to-noise ratio (PSNR) or structural similarity (SSIM)~\cite{wang2004image}, which measures structural difference between two images in factors such as brightness and contrast. But in practice, they are found not to match human perception very well~\cite{liu2013no}. A new metric PieAPP~\cite{prashnani2018pieapp} is proposed recently to measure the \textit{human perceptual} distance between two images. But PieAPP might be a little complicated, as the metric itself is produced by a neural network.

In many other vision tasks, it seems that the preparation of the dataset has nothing to do with intelligence, but only depends on sufficient labor force. However, how to prepare the dataset does become a non-trivial problem for deblurring.

The majority of traditional methods~\cite{levin2009understanding} simply use synthetic datasets with uniformly blurred images generated by convolving a sharp image with a known blue kernel. While synthesis can generate a great number of blurred/unblurred pairs very rapidly, the way that images are blurred (2D, spatially invariant) is far away from that in the real world (6D, spatially variant).

\begin{figure}[t]
\centering
\includegraphics[height=5.5cm]{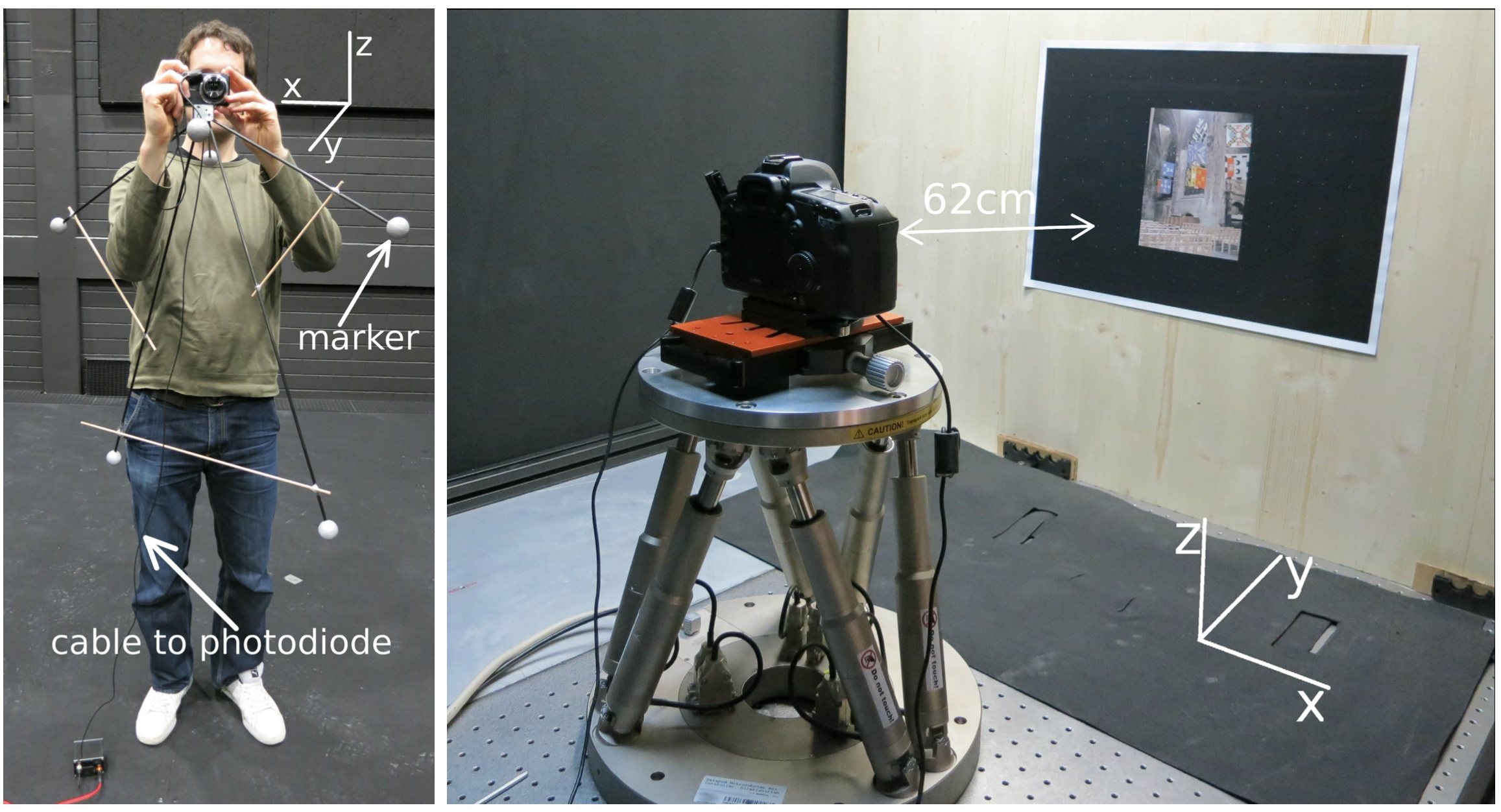}
\caption{(Left) Light-weight structure with reflective spherical markers for recording camera shake with a Vicon system. (Right) Camera attached to a high-precision hexapod robot to take pictures with played back camera shakes in a controlled setting. Source: K{\"o}hler \textit{et al.}~\cite{kohler2012recording}.}
\label{6D_shake}
\end{figure}

K{\"o}hler \textit{et al.}~\cite{kohler2012recording} designs a laboratory environment that can generates 6D camera shakes and royally records the blurring process. Fig. \ref{6D_shake} shows the settings of their equipment. However, blurry images are constrained to a laboratory setting, and it is obviously very time-consuming the enlarge the size of the dataset. In fact, this dataset only consists of 4 latent images and 12 different blurring types for each of them.

Lai \textit{et al.}~\cite{lai2016comparative} generates a synthetic dataset by first recording some 6D camera trajectories and then applying them to images both uniformly and non-uniformly. Some blurry images `in the wild' are either captured in person or collected from the internet. They are also added to the dataset. However, blurry images in the wild naturally lack corresponding sharp images, while synthetic images are reported~\cite{nah2017deep} to be not aligned well with sharp latent images from the perspective of PSNR and SSIM.

So far, the most elaborate and successful dataset is arguably the GOPRO dataset present by Nah \textit{et al.}~\cite{nah2017deep}. The GOPRO dataset adopts a so-called `averaging method' to synthesize blurry images, which originates from the simulation of the very nature that leads to motion blur.

When the shutter button is pressed, camera sensor starts to receive light signal. During the exposure process, sharp image stimulation is accumulated, generating blurry images. The integrated signal is finally transformed into pixel value by a nonlinear camera response function (CRF)~\cite{tai2013nonlinear}. Mathematically, the blur accumulation can be modeled as
\begin{align}
    B=g(\frac{1}{T}\int_{t=0}^TS(t)\mathrm{d}t)\simeq g(\frac{1}{M}\sum_{i=0}^{M-1}S[i]),
\end{align}
where $B$ is the blurry image, $g$ is the CRF that maps a latent signal into an observed image, $T$ is the exposure time, $S(t)$ is the latent sharp image at time $t$, $M$ the number of sampled points, and $S[i]$ is the $i$-th sharp frame captured during the exposure process.

\begin{figure}[t]
\centering
\includegraphics[height=3.3cm]{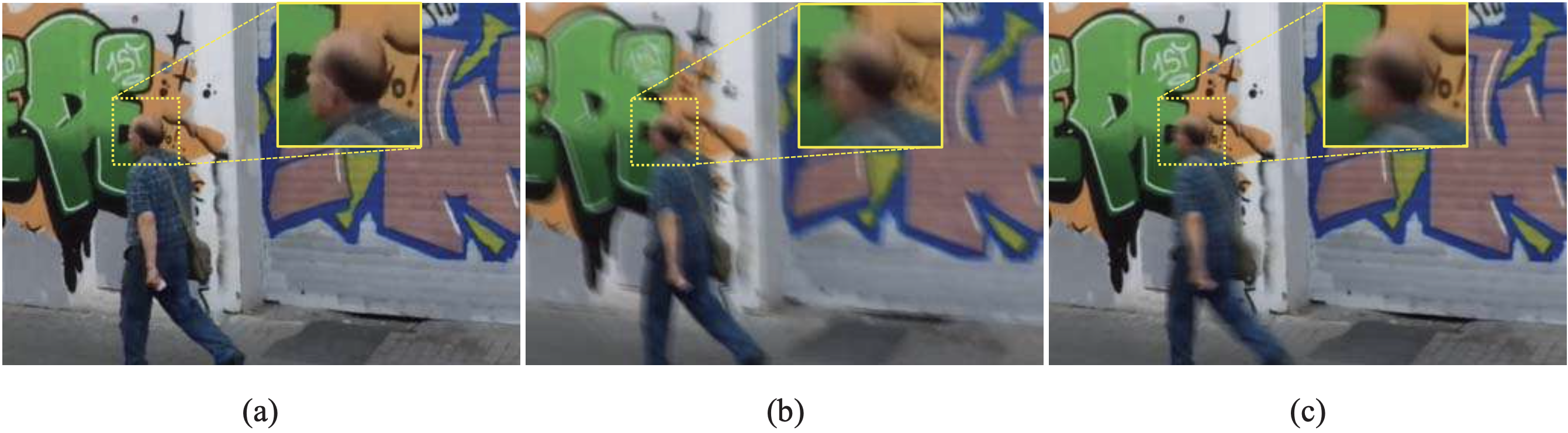}
\caption{(a) Ground truth sharp image. (b) Blurry image generated by convolving a uniform blur kernel. (c) Blurry image generated by the averaging method. In (b), both the foreground person and the background is blurry. In comparison, the background in (c) remains sharp, which is more reasonable in most cases. Source: Nah \textit{et al.}~\cite{nah2017deep}.}
\label{gopro}
\end{figure}

In practice, Nah \textit{et al.} took 240fps videos with a GOPRO camera and then averaging varying number (7-13) of successive latent sharp frames to produce blurs of different strengths. The mid-frame among the successive sharp frames is selected as the corresponding sharp image. Fig.~\ref{gopro} exhibits the effect of the averaging method.

\subsection{Optimization-based Methods}
As is stated in Section~\ref{section_problem}, blind deblurring is an ill-posed problem. In order to select a pair of good $(x,k)$ from infinite candidates, researchers have developed tricks, priors and additional information of various kinds. While technical details are diverse, most optimization-based methods follow a maximum a posteriori (MAP) estimation framework to enable the optimization process. In general, an MAP framework can be formalized as
\begin{align}
    (x^*,k^*)=\argmax_{x,k}p(x,k|y)p(x)p(k).
\end{align}
In practice, the framework is often expressed as
\begin{align}
    (x^*,k^*)=\argmin_{x,k}\|x\otimes k-y\|+\lambda\varphi(x)+\mu\psi(k),
\end{align}
where $\lambda$ and $\mu$ are weights, and $\varphi(x)$ and $\psi(k)$ are priors on the latent image and the kernel, respectively. Now we are going to introduce some of the priors proposed in literature.

%on the sharp image $x$~\cite{chan1998total,fergus2006removing,krishnan2011blind,sun2013edge,pan2016l_0,zuo2016learning} and the kernel $k$~\cite{liu2014blind,michaeli2014blind,pan2017deblurring,ren2016image,yan2017image}. 

\begin{figure}[t]
\centering
\includegraphics[height=4.3cm]{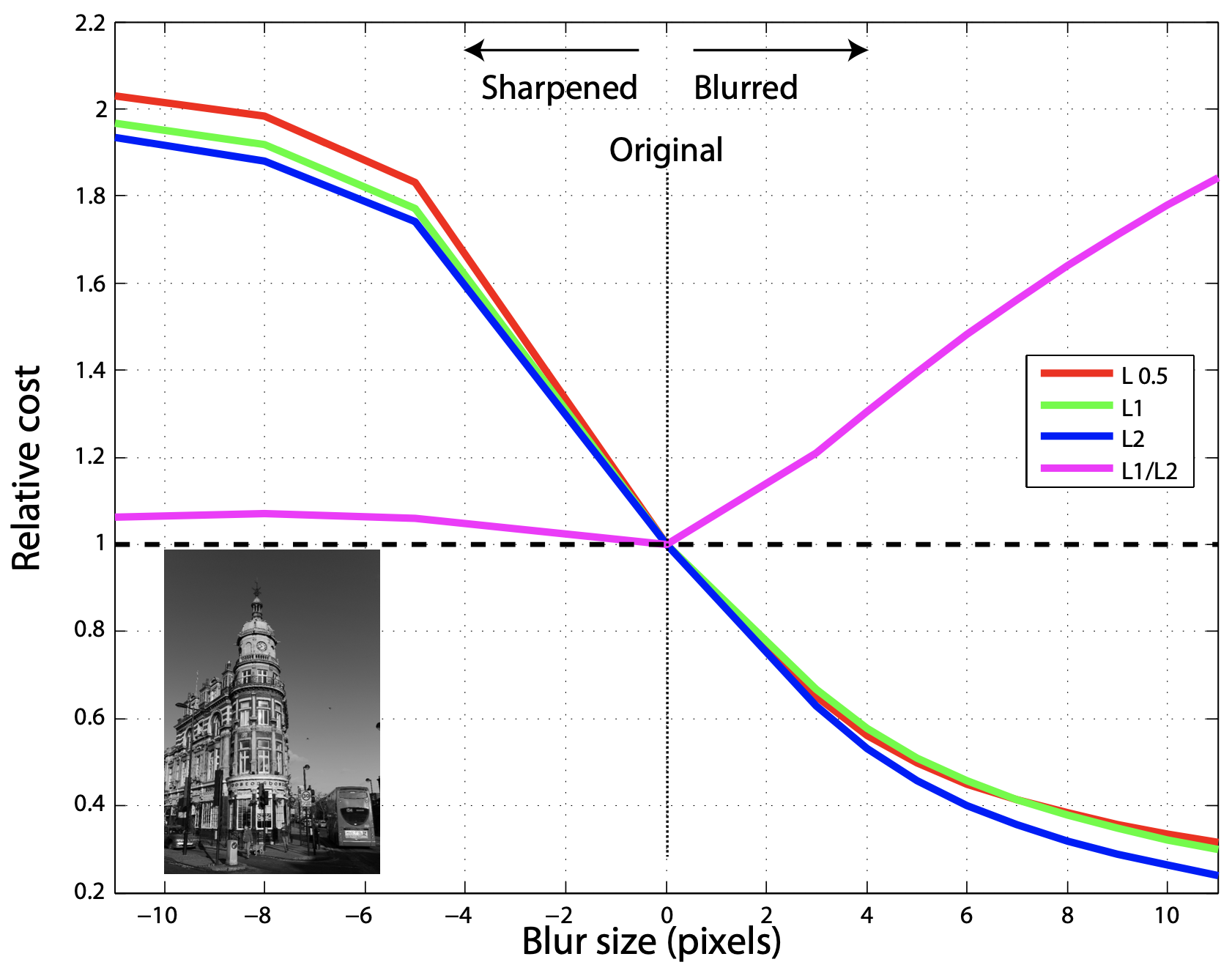}
\includegraphics[height=4.3cm]{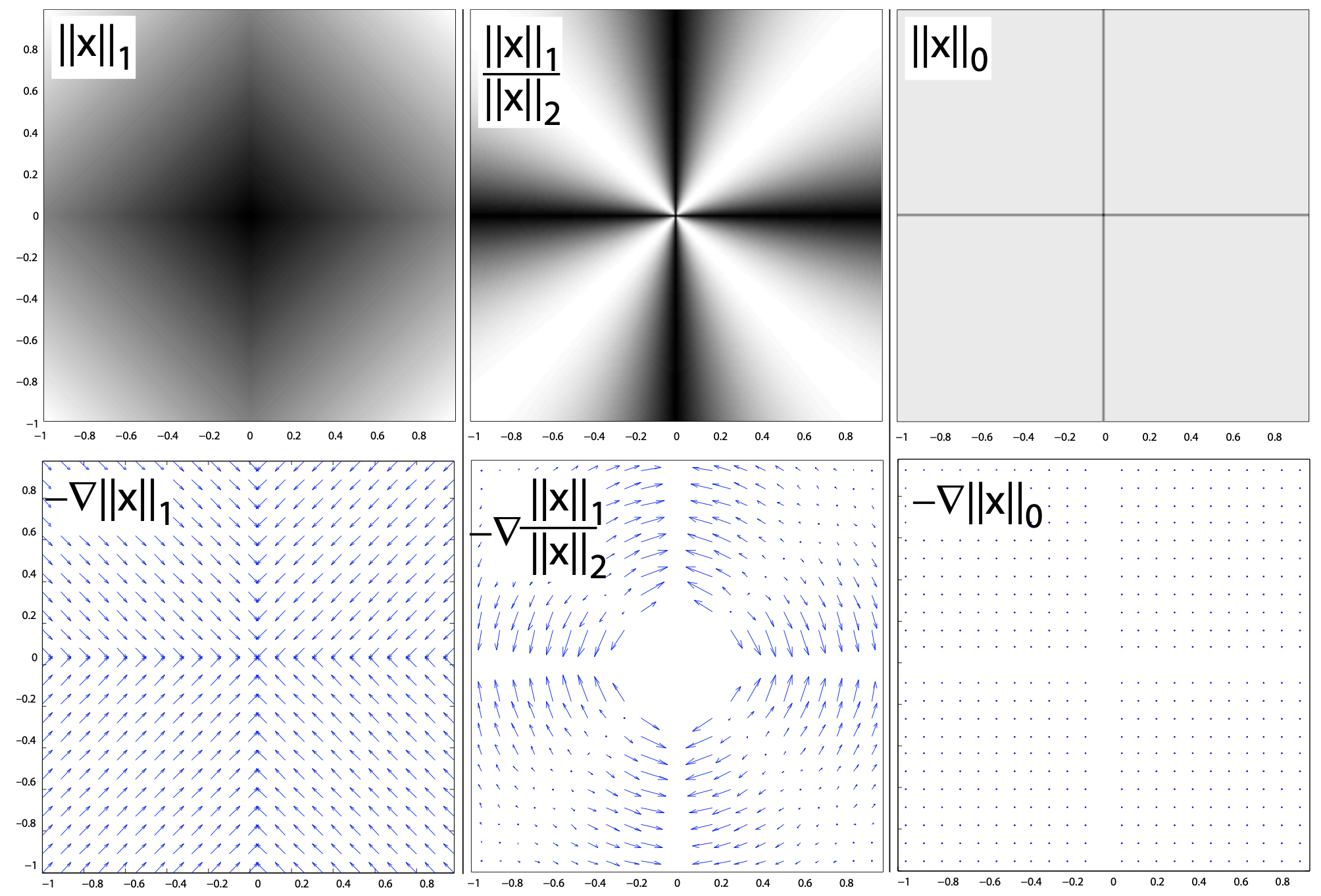}
\caption{(Left) A comparison of the $l_1/l_2$ regularizer to other approaches. (Right) A visualization of $l_1$, $l_1/l_2$ and $l_0$ functions (top row) and their negative gradient fields (bottom row). Source: Krishnan \textit{et al.}~\cite{krishnan2011blind}}
\label{l1/l2}
\end{figure}

A considerable number of existing methods~\cite{chan1998total,fergus2006removing,levin2009understanding,levin2011efficient,krishnan2011blind,xu2013unnatural,pan2014deblurring,pan2016l_0,perrone2014total} take advantage of statistical priors to help estimate the blur kernel. To be more specific, the gradients of natural images should be concentrated at zero and be sparse at large values, as they are found to follow a heavy-tailed distribution~\cite{field1994goal,roth2005fields}. In the seminal work of~\cite{chan1998total}, Chan and Wong introduce total variational (TV) regularization, which is a common regularization in denoising methods that can suppress image gradients, to both the latent image and the kernel in blind deconvolution. Fergus \textit{et al.}~\cite{fergus2006removing} leverage zero-mean Mixture of Gaussian to fit the heavy-tailed prior. A variational Bayesian framework is employed to estimate the kernel. The theory of Levin \textit{et al.} \cite{levin2009understanding} emphasizes the strength of estimating the kernel alone over estimating both the image and the kernel. The experiments of \cite{levin2009understanding} claims Fergus \textit{et al.}~\cite{fergus2006removing} outperforms other then-existing methods. Based on Fergus \textit{et al.}~\cite{fergus2006removing}, Levin \textit{et al.}~\cite{levin2011efficient} provide a significantly simpler derivation via the expectation-maximization (EM) approach. Krishnan \textit{et al.}~\cite{krishnan2011blind} point out the ratio of the $l_1$ norm to the $l_2$ norm of the gradients can clearly signify the difference between blurry images and sharp images (See Fig.~\ref{l1/l2}). This observation makes the deblurring algorithm fast and easy to complement. While the theory of Levin \textit{et al.} \cite{levin2009understanding} favors optimization on the kernel alone, numerous successful methods empirically verify the feasibility of the very opposite way, which seems to be a paradox. Perrone and Favaro~\cite{perrone2014total} find out that both the theory and the empirical evidence are correct. They also claim that details in implementation can dramatically decide the convergence of the algorithm.

\begin{figure}[b]
\centering
\includegraphics[height=1.85cm]{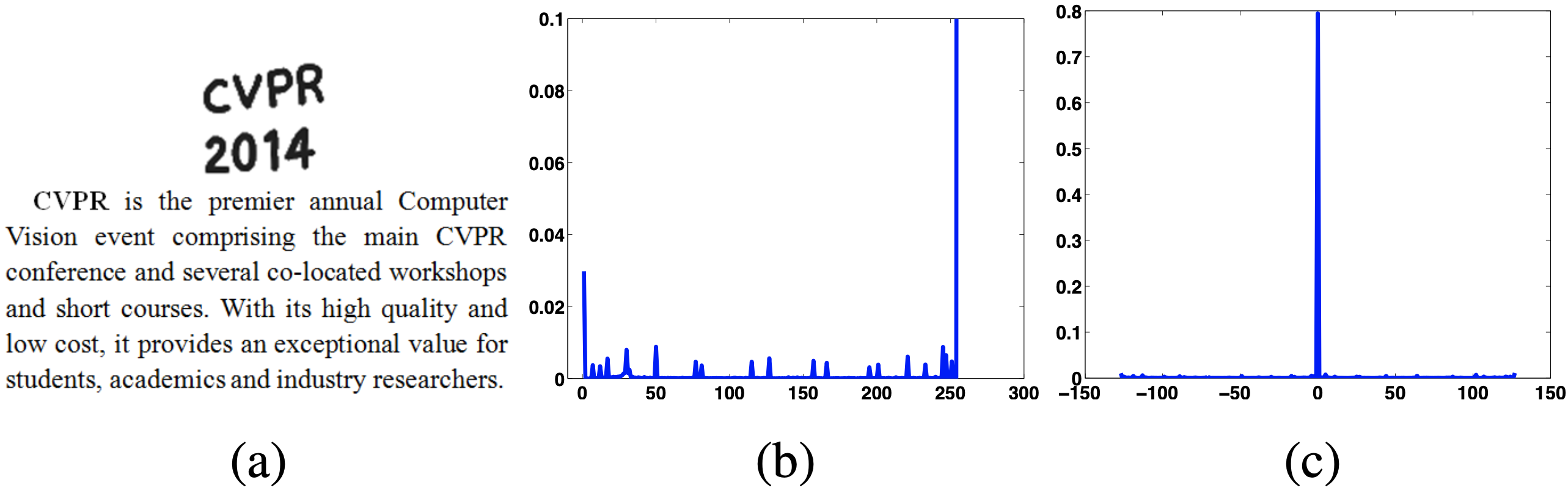}
\includegraphics[height=1.85cm]{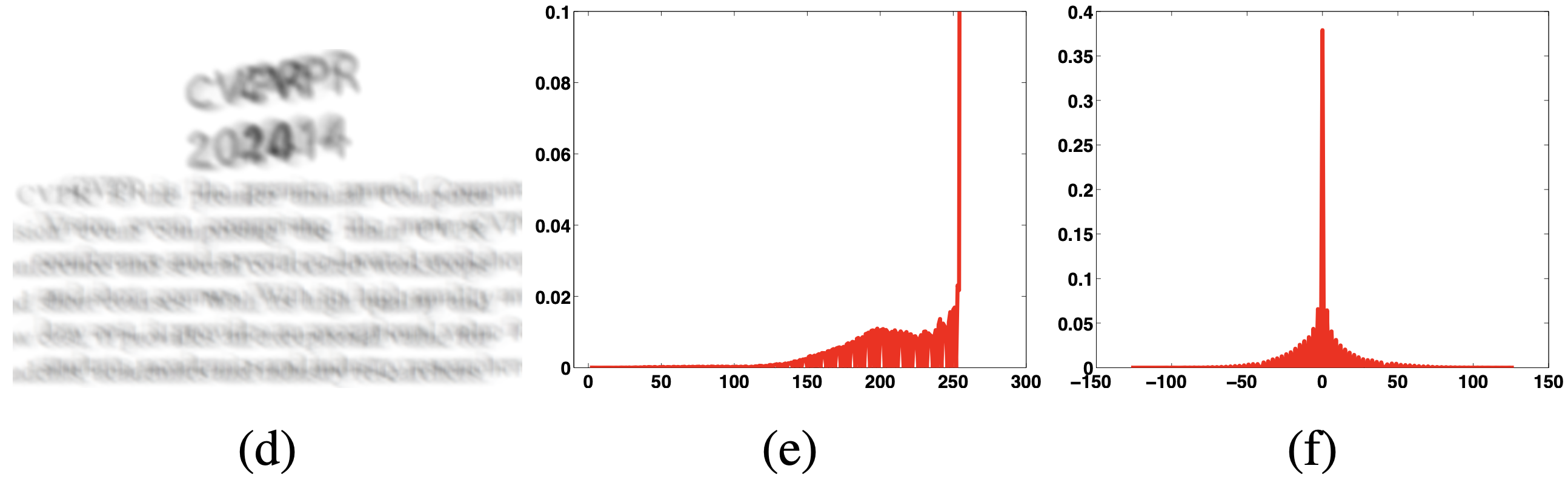}
\caption{Statistics of text images. (a) clean text image. (b) histogram of pixel intensities from (a). (c) histogram of horizontal gradients from (a). (d) blurred image. (e) histogram of pixel intensities from (d). (f) histogram of horizontal gradients from (d). Source: Pan \textit{et al.}~\cite{pan2014deblurring}}
\label{text_prior}
\end{figure}

On the other hand, Xu \textit{et al.}~\cite{xu2013unnatural} argue that the success of previous MAP based methods stems from their respective intermediate steps, which implicitly or explicitly create an unnatural representation containing high-contrast and step-like salient image structures. To utilize the properties of step-like edges, Xu \textit{et al.} propose an $l_0$ sparse approximation scheme which is more general and more likely to converge. Likewise, Pan \textit{et al.}~\cite{pan2014deblurring} utilize the $l_0$ regularized prior to deblur text images. Due to the statistical properties of text images (See Fig.~\ref{text_prior}), Pan \textit{et al.} apply the $l_0$ prior not only to the gradient, but also to the intensity. A follow-up article~\cite{pan2016l_0} suggests this method can be extended to natural images deblurring and non-uniform deblurring.

\begin{figure}[t]
\centering
\includegraphics[height=4cm]{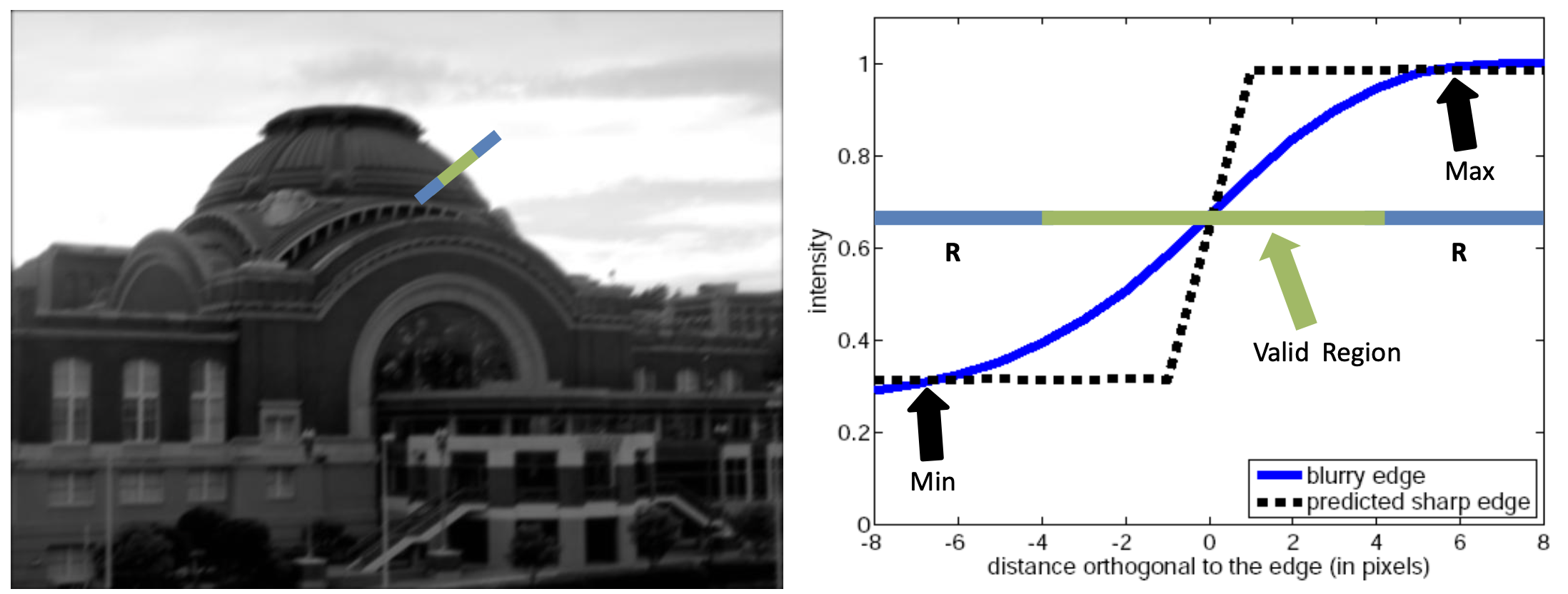}
\caption{(Left) A blurry image and an example 1D profile normal to an edge. (Right) Blue line is the observed blurry edge. Dashed black line is the predicted sharp edge. Source: Joshi \textit{et al.}~\cite{joshi2008psf}}
\label{edge}
\end{figure}

\begin{figure}[b]
\centering
\includegraphics[height=2cm]{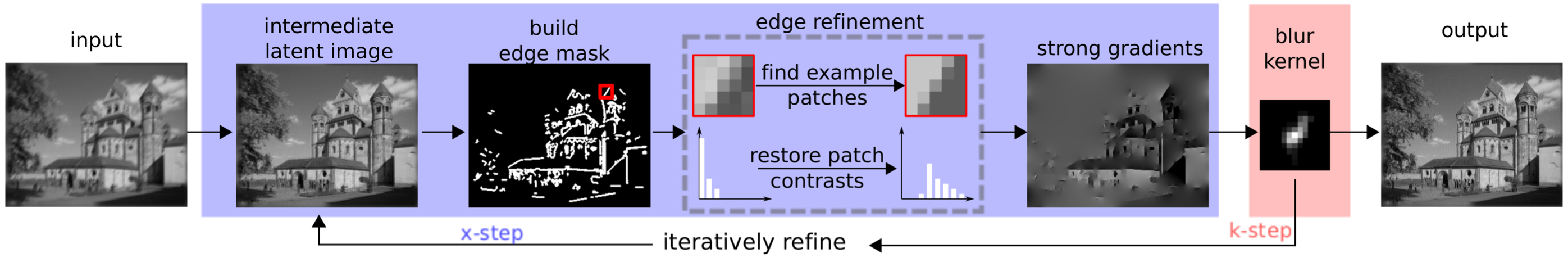}
\caption{The pipeline of an iterative patch-matching method~\cite{sun2013edge}. The \textit{edge refinement} procedure enforces edges to become sharp and increases local contrast for edge patches. Source: Sun \textit{et al.}~\cite{sun2013edge}}
\label{patch}
\end{figure}

Apart from sparsity priors, many methods~\cite{joshi2008psf,cho2009fast,xu2010two,sun2013edge} take advantage of sharp edges detection to estimate the blur kernel. Joshi \textit{et al.}~\cite{joshi2008psf} predict a sharp edge by propagating the max and min values along the profile normal to observed edges detected by classic methods. The algorithm then leverages the difference between the predicted and observed values along the edges to solve for spatially variant blur kernels. Fig.~\ref{edge} shows the propagation process. Similar to Joshi \textit{et al.}, Cho and Lee~\cite{cho2009fast} accelerate the iterative estimation process by utilizing strong edges for image restoration and derivatives for kernel estimation. Additionally, a multi-scale framework is added to their method to help process larger blurs in a coarse-to-fine fashion. Xu and Jia~\cite{xu2010two} argue that not all strong edges profit kernel estimation and propose a new metric to measure the usefulness of image edges. They also achieve efficient and high-quality kernel estimation via iterative support detection (ISD) kernel refinement which can enforce the sparsity prior. In~\cite{sun2013edge}, Sun \textit{et al.} state that heuristic filters such as shock and bilateral filtering which previous methods favor are often unstable. Therefore, they derive a patch-matching strategy to improve the performance of edge-based methods. The so-called `patches' are atomic elements that form the structural part of the image, namely, edges, corners, T-junctions, \textit{etc}. Fig.~\ref{patch} shows the pipeline of this method. Surprisingly, it is found that simple synthetic patch prior can generate even better results than learned statistical prior.

\begin{figure}[t]
\centering
\includegraphics[height=3.8cm]{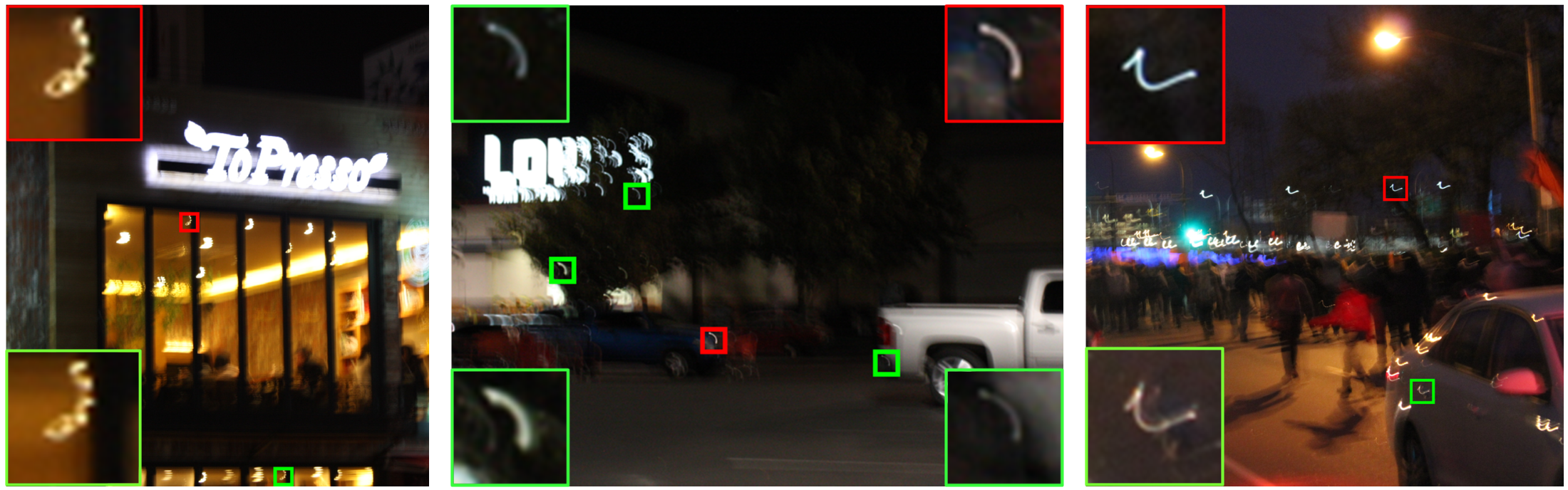}
\caption{Examples of light streak detection. The red box indicates the best light streak patch and the green boxes show additional light streak patches that are automatically identified. Source: Hu \textit{et al.}~\cite{hu2014deblurring}}
\label{light_streak}
\end{figure}

In fact, besides sparsity and edge-based prior, there are many other novel priors~\cite{jia2007single,shan2008high,hu2014deblurring,cao2015scene} or additional information~\cite{ben2003motion,yuan2007image,hu2016image} leveraged to help boost the performance of deblurring algorithms. Ben-Ezra and Nayar~\cite{ben2003motion} build a hybrid imaging system by assembling a high-resolution camera and a low-resolution DV together. Yuan \textit{et al.}~\cite{yuan2007image} manage to enhance the quality of the image via a different hybrid imaging scheme. The final image is synthesized by a pair of blurry image taken with long shutter time and low ISO and noisy image taken with short shutter time and high ISO. Jia~\cite{jia2007single} takes advantage of transparency on the boundary of a moving object to recover sharp images. Shan \textit{et al.}~\cite{shan2008high} take advantage of local gradient prior on homogeneous areas to suppress ringing artifacts. Hu \textit{et al.}~\cite{hu2014deblurring} notice that the light streaks (See Fig.~\ref{light_streak}) could contain rich blur information. They design methods to automatically detect useful light streaks and significantly improve the quality of challenging low-light images. Cao \textit{et al.}~\cite{cao2015scene} suggest texts in natural images can be efficiently identified and then be used to recover a sharp natural scene. Hu \textit{et al.} suggest that low-resolution inertial sensors in smartphones, \textit{i.e.}, gyroscopes and accelerometers can be adopted to reconstruct spatially-variant PSFs for image deblurring.

As is mentioned, there lacks a solve-for-all metric for the deblurring problem. Therefore, comparative studies~\cite{levin2009understanding,kohler2012recording,lai2016comparative} raise different evaluation methods that focus on different aspects. Among them, the human subjective study conducted by Lai \textit{et al.}~\cite{lai2016comparative} seems to be the most thoughtful method. Most studies agree that Xu and Jia~\cite{xu2010two} is a good-performer, especially in uniform deblurring settings. Pan \textit{et al.}~\cite{pan2014deblurring} outperforms other methods in real and non-uniform settings. Xu \textit{et al.}~\cite{xu2013unnatural}, Sun \textit{et al.}~\cite{sun2013edge}, and Krishnan \textit{et al.}~\cite{krishnan2011blind} are also accepted to be good deblurring methods.

\begin{figure}[t]
\centering
\includegraphics[height=3.6cm]{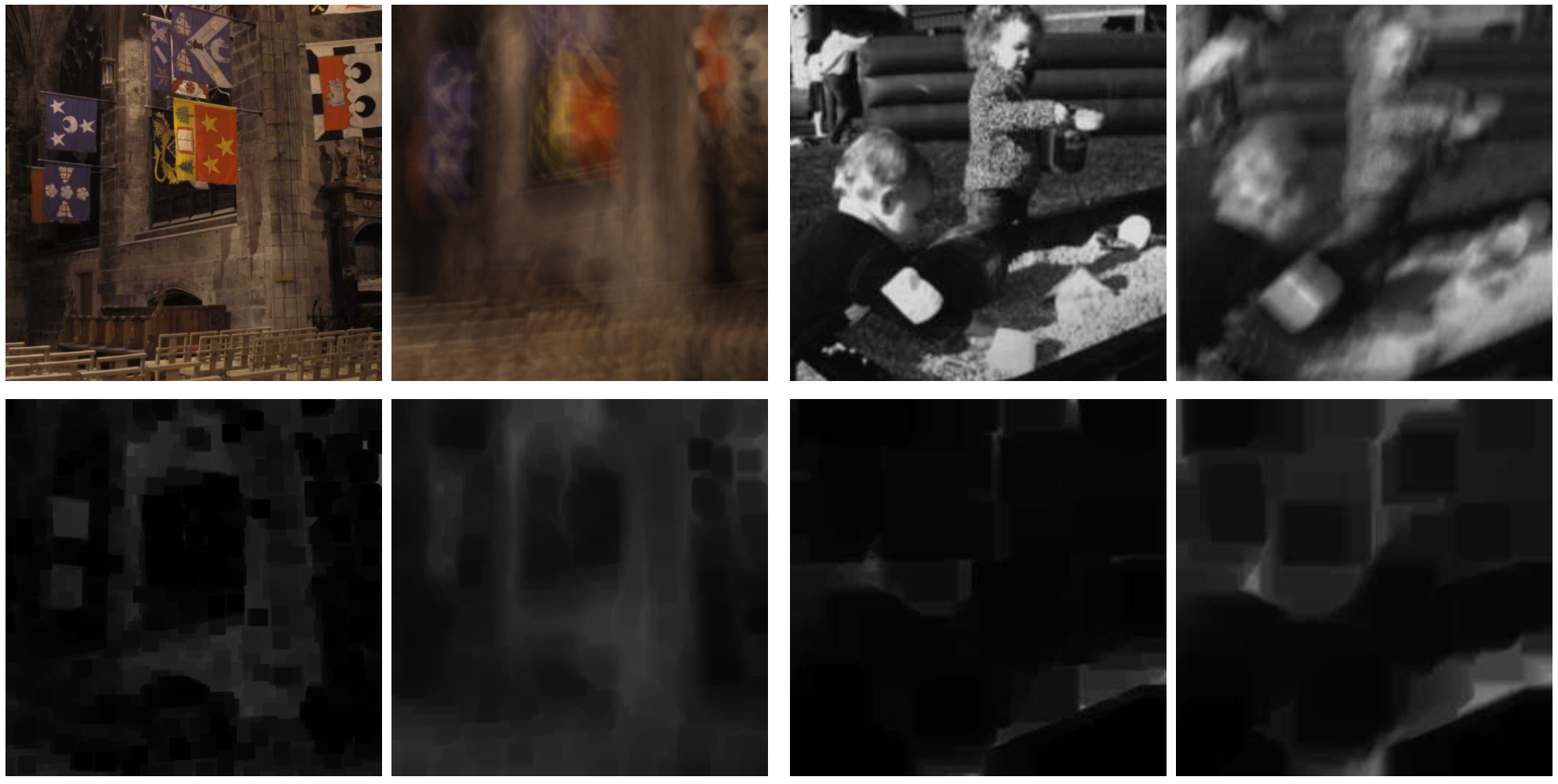}
\includegraphics[height=3.6cm]{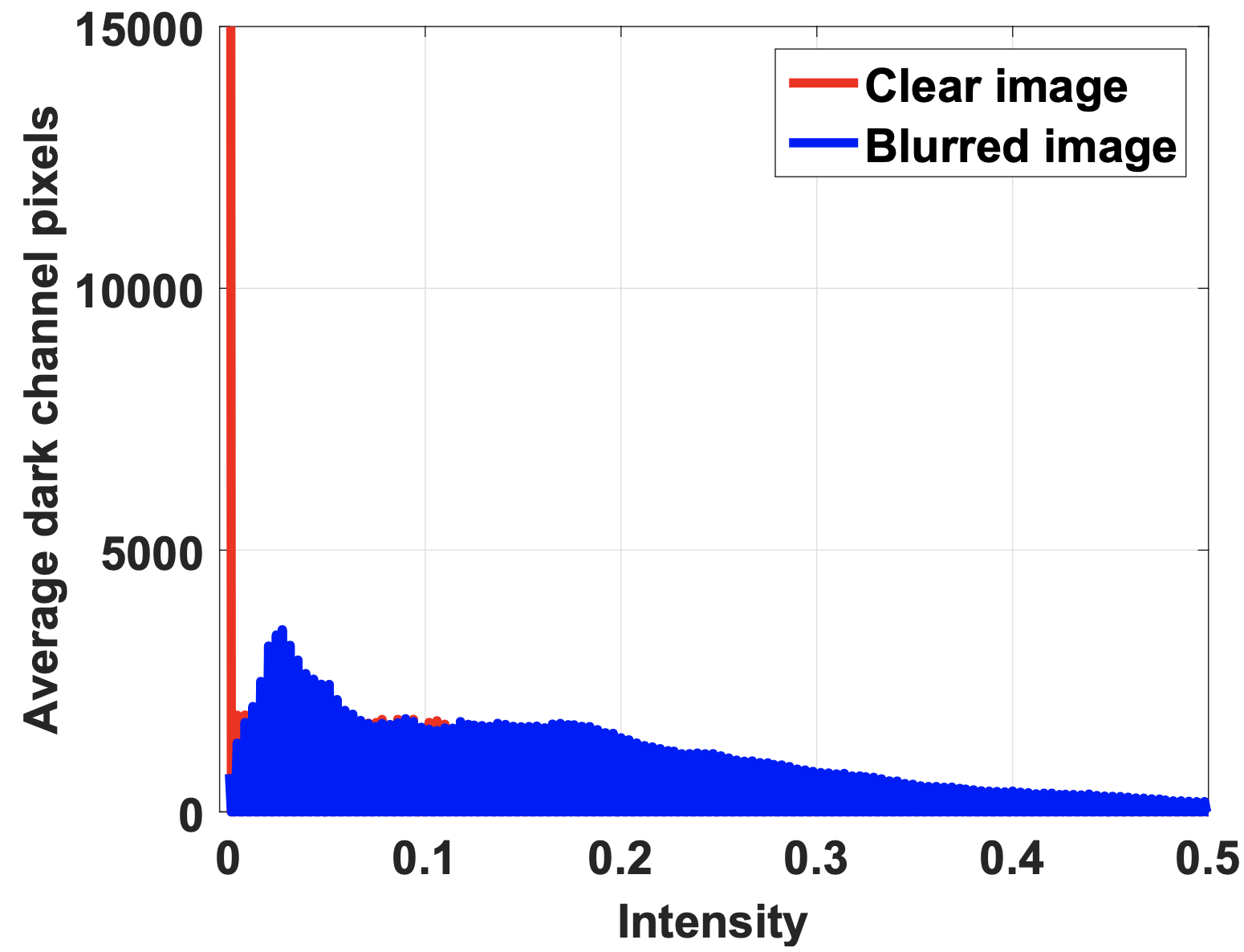}
\caption{(Left) Blurry images have fewer dark channels than clear images. Top row: images; bottom row: corresponding dark channels computed with an image patch size of 35×35. (Right) Intensity histograms for dark channels of clear (red) and blurry (blue) images. Source: Pan \textit{et al.}~\cite{pan2016blind}}
\label{dark}
\end{figure}

Recently, researchers~\cite{pan2016blind,yan2017image} start to take notice of a new type of prior, \textit{i.e.}, the dark channel prior and its variants. The dark channel prior is proposed in the famous haze removal work by He \textit{et al.}~\cite{he2010single}, based on the observation that for most of the non-sky patches of a haze-free outdoor images, at least one color channel has very low intensity at some pixels. Formally, for an image $I$, the dark channel $D(I)$ is given by
\begin{align}
    D(I)(x)=\min_{y\in\mathcal{N}(x)}\min_{c\in {r,g,b}}I^c(y),
\end{align}
where x and y denote pixel locations, $\mathcal{N}(x)$ is an image patch centered at $x$, and $I^c$ is the $c$-th color channel. Pan \textit{et al.}~\cite{pan2016blind} discover that this prior is also useful to distinguish sharp images from blurry ones (See Fig.~\ref{dark}) and adapt it to the deblurring problem. Based on Pan \textit{et al.}~\cite{pan2016blind}, Yan \textit{et al.}~\cite{yan2017image} propose a symmetric bright channel prior, and manage to leverage both of the priors, which they name as extreme channels prior. Despite the lack of third-party evaluations, qualitative results indicate Yan \textit{et al.}~\cite{yan2017image} is potentially the state-of-the-art method for optimization-based blind deblurring.

\subsection{Learning-based Methods}
Neural networks have been successfully applied to low-level vision tasks such as denoising~\cite{jain2008natural} and super-resolution~\cite{dong2014learning}. Not surprisingly, quite a few learning-based methods are raised as well to deal with the similar but in many senses more challenging blind deblurring problem.

Following the custom of optimization-based methods, early learning-based methods utilize neural networks for kernel estimation. Recall the patch-matching technique in \cite{sun2013edge}. Atomic graphic elements in patch candidates easily remind AI researchers of the visualization results of `what is learned' in the first few layers of a CNN. Probably inspired by this observation, Sun \textit{et al.}~\cite{sun2015learning} use a CNN to predict the probabilistic distribution of motion blur at the patch level. A Markov random field model is then used to infer a dense non-uniform motion blur field enforcing motion smoothness. Gong \textit{et al.}~\cite{gong2017motion} propose a natural development of this approach by eliminating the time-consuming patch-wise classification process. Instead, the CNN directly outputs pixel-level motion vector field. To imitate the procedures of traditional MAP-based methods, Schuler \textit{et al.}~\cite{schuler2015learning} manage to represent feature extraction, kernel estimation and image estimation modules with a trainable deep neural network. These modules are stacked to simulate multiple iterations in traditional methods. Chakrabarti \textit{et al.}~\cite{chakrabarti2016neural} propose a more aggressive approach where a neural network directly infers the Fourier coefficients of the inverse filter.

The pioneering work of Nah \textit{et al.}~\cite{nah2017deep} introduce a CNN that directly restores a latent sharp image without kernel estimation, as an image-to-image framework can avoid errors induced by the estimated blur kernel. Its multi-scale architecture  enables it to follow a coarse-to-fine manner, which proves to be helpful in traditional methods. Tao \textit{et al.}~\cite{tao2018scale} propose a scale-recurrent network where CNNs responsible for different scales share parameters. This scheme reduces the number of parameters and allows the network to capture more information. Kupyn \textit{et al.}~\cite{kupyn2018deblurgan} introduce a training method based on a GAN, which can effectively reconstruct sharp edges. The network consists of an encoder–decoder type CNN and a generator with residual layers. Zhang \textit{et al.}~\cite{zhang2018dynamic} argue that the nature of the blurring process asks for long-range spatial dependencies, but the receptive fields of CNNs are usually not large enough. Based on this observation, they propose a hybrid network that employs a recurrent neural network (RNN) as well as a CNN.

\begin{comment}
\section{Highlights of Classical Non-blind Methods}
Wiener filter~\cite{gonzalez2008digital} and Richardson-Lucy (RL) method~\cite{richardson1972bayesian,lucy1974iterative} are two well-known classical methods for non-blind deconvolution. Though invented long ago, they are very inspiring and widely adopted. From the perspective of Bayesian inference, Wiener filter adopts MAP and a Gaussian prior on the latent image, while RL is an iterative methods that converges at the maximum likelihood solution. Thus, before exploring more complicated contemporary methods, we first introduce the two classical non-blind methods.

\subsection{Wiener deconvolution}

\subsection{The Richardson-Lucy method}

\section{Highlights of Levin \textit{et al.}~\cite{levin2009understanding}}
\label{section_theory_bayesian}
\end{comment}

\clearpage
\bibliographystyle{splncs04}
\bibliography{egbib}
\end{document}